\pdfoutput=1

\documentclass[11pt]{article}

\usepackage{ACL2023}

\usepackage{times}
\usepackage{latexsym}

\usepackage[T1]{fontenc}

\usepackage[utf8]{inputenc}

\usepackage{microtype}

\usepackage{inconsolata}

\usepackage{amsmath,amsfonts}
\usepackage{booktabs,arydshln}
\usepackage{multirow}
\usepackage{ulem}
\usepackage{pifont}
\usepackage{relsize}
\usepackage{enumitem}
\usepackage{graphicx}
\usepackage{tikz}

\makeatletter
\def\adl@drawiv#1#2#3{%
        \hskip.5\tabcolsep
        \xleaders#3{#2.5\@tempdimb #1{1}#2.5\@tempdimb}%
                #2\z@ plus1fil minus1fil\relax
        \hskip.5\tabcolsep}
\newcommand{\cdashlinelr}[1]{%
  \noalign{\vskip\aboverulesep
           \global\let\@dashdrawstore\adl@draw
           \global\let\adl@draw\adl@drawiv}
  \cdashline{#1}
  \noalign{\global\let\adl@draw\@dashdrawstore
           \vskip\belowrulesep}}
\makeatother

\newcommand{\cmmnt}[1]{\ignorespaces}
\newcommand{\trm}[1]{\textrm{#1}}

\newcommand{\symbolsecref}[1]{($\S$~\ref{#1})}

\usepackage{breakurl}
\usepackage[breaklinks]{hyperref}
\newcommand{\grayit}[1]{\textcolor{gray}{#1}}

\newcommand{\roberta}[0]{RoBERTa}
\newcommand{\houlsby}[0]{Houlsby}
\newcommand{\pfeiffer}[0]{Pfeiffer}

\newcommand{\spartan}[0]{\textsc{spartan}}

\newcommand{\cola}[0]{CoLA}
\newcommand{\sst}[0]{SST}
\newcommand{\mrpc}[0]{MRPC}
\newcommand{\stsb}[0]{STSB}
\newcommand{\qqp}[0]{QQP}
\newcommand{\mnli}[0]{MNLI}
\newcommand{\qnli}[0]{QNLI}
\newcommand{\rte}[0]{RTE}
\newcommand{\wnli}[0]{WNLI}

\newcommand{\raspi}[0]{Ras-Pi}
\newcommand{\raspifull}[0]{Raspberry Pi 4}

\newcommand{\iphone}[0]{iPhone}
\newcommand{\matstorage}[0]{1.1$\times$}
\newcommand{\houlsbystorage}[0]{1.2$\times$}
\newcommand{\robertastorage}[0]{9$\times$}

\newcommand{\numparents}[0]{$N_{p}$}
\newcommand{\numchild}[0]{$N_{c}$}
\newcommand{\inputvector}[0]{$v_{I}$}
\newcommand{\outputvector}[0]{$v_{O}$}
\newcommand{\topk}[0]{top-$K$}
\newcommand{\parentind}[0]{$P_{\trm{~IND}}$}
\newcommand{\parentprob}[0]{$g_{\trm{parent}}$}

\title{SPARTAN: Sparse Hierarchical Memory for Parameter-Efficient Transformers}

\author{
Ameet Deshpande\textsuperscript{1}, Md Arafat Sultan\textsuperscript{2}, Anthony Ferritto\textsuperscript{3}, Ashwin Kalyan\textsuperscript{4},\\
\textbf{Karthik Narasimhan\textsuperscript{1}, Avirup Sil}\textsuperscript{2}\\
Princeton University\textsuperscript{1}, IBM Research AI\textsuperscript{2}, Amazon AWS AI\textsuperscript{3}, The Allen Institute for AI\textsuperscript{4}\\
\texttt{asd@cs.princeton.edu}, \texttt{arafat.sultan@ibm.com}, \texttt{ferritta@amazon.com},\\
\texttt{ashwinkv@allenai.org}, \texttt{karthikn@cs.princeton.edu}, \texttt{avi@us.ibm.com}
}

\begin{document}
\maketitle

\tikz[remember picture,overlay]
\path (current page.north east) ++ (-1.5,-1.5) node {\includegraphics[width=13mm]{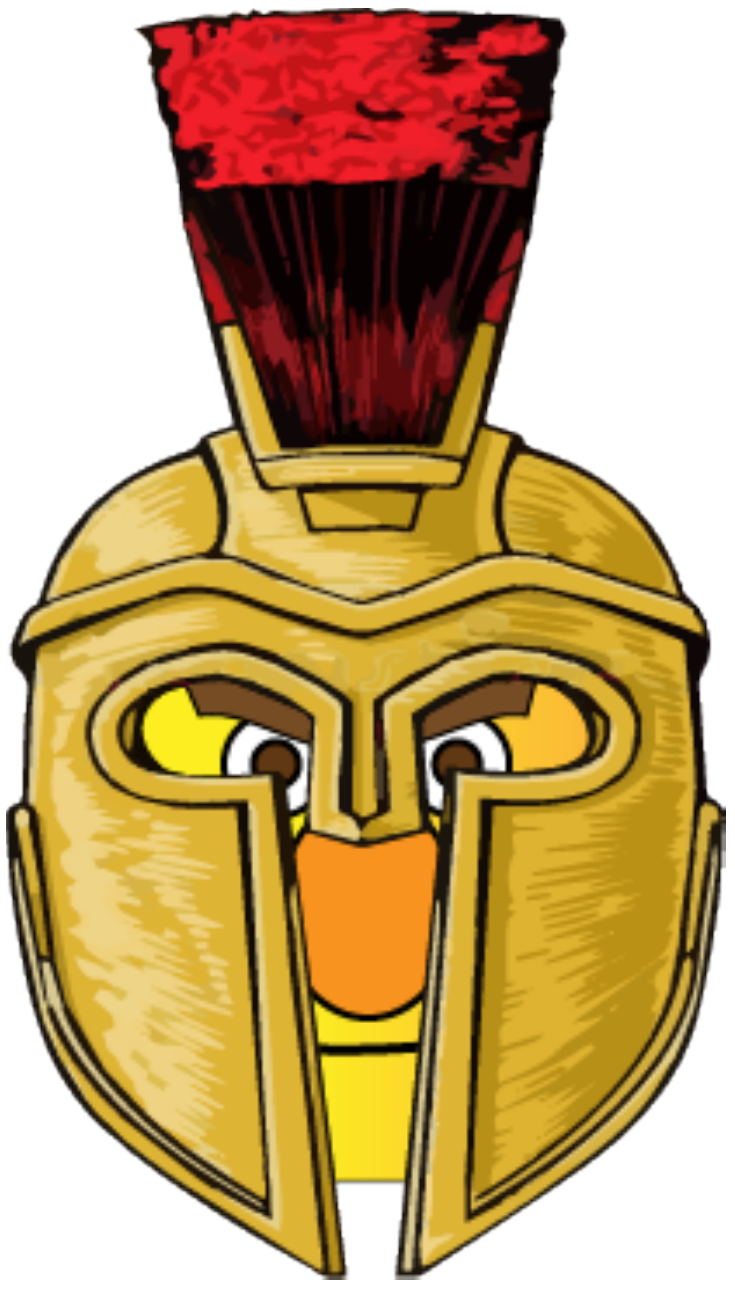}};

\begin{abstract}

Fine-tuning pre-trained language models (PLMs) achieves impressive performance on a range of downstream tasks, and their sizes have consequently been getting bigger.
Since a different copy of the model is required for each task, this paradigm is infeasible for storage-constrained edge devices like mobile phones.
In this paper, we propose \spartan{}, a parameter efficient (PE) and computationally fast architecture for edge devices that adds hierarchically organized sparse memory after each Transformer layer.
\spartan{} freezes the PLM parameters and fine-tunes only its memory, thus significantly reducing storage costs by re-using the PLM backbone for different tasks.
\spartan{} contains two levels of memory, with only a sparse subset of parents being chosen in the first level for each input, and children cells corresponding to those parents being used to compute an output representation.
This sparsity combined with other architecture optimizations improves \spartan{}'s throughput by over $90\%$ during inference on a Raspberry Pi 4 when compared to PE baselines (adapters) while also outperforming the latter by $0.1$ points on the GLUE benchmark.
Further, it can be \textit{trained} $34\%$ faster in a few-shot setting, while performing within $0.9$ points of adapters.
Qualitative analysis shows that different parent cells in \spartan{} specialize in different topics, thus dividing responsibility efficiently.
\footnote{Code: \url{https://github.com/princeton-nlp/SPARTAN}}

\end{abstract}

\section{Introduction}
\label{sec:intro}

\definecolor{parent}{HTML}{c4b0d5}
\definecolor{children}{HTML}{8ed3f8}
\definecolor{input}{HTML}{ffb2aa}
\definecolor{output}{HTML}{b5eb95}

\begin{figure}[t]
    \centering
    \includegraphics[width=\columnwidth]{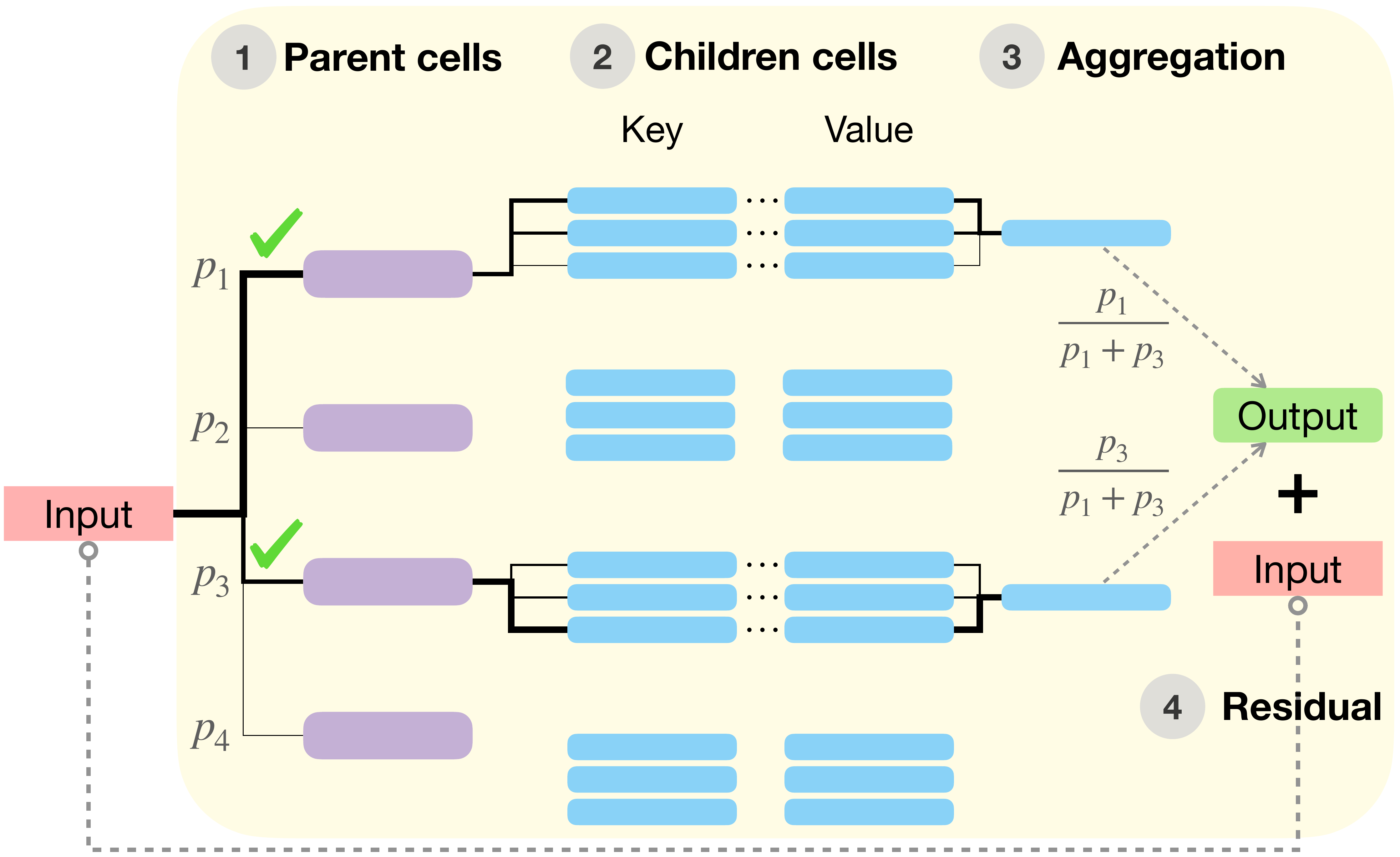}
    \caption{
        \spartan{} adds a sparse, hierarchically organized memory after each Transformer layer that is shared by all positions in the input sequence.
        \textcircled{\scriptsize 1} The \colorbox{input}{input} corresponds to a single position, and chooses a sparse subset of \colorbox{parent}{parent cells} based on a computed probability distribution (here, top-$2$).
        \textcircled{\scriptsize 2} The corresponding \colorbox{children}{children cells} are used to compute an input-conditioned representation,
        \textcircled{\scriptsize 3} which is aggregated via a weighted sum based on the probability distribution in step 1 to give the \colorbox{output}{output}.
        \textcircled{\scriptsize 4} It is added to the \colorbox{input}{input} through a residual connection which serves as the input to the next Transformer layer.
        \spartan{} outperforms baselines on GLUE while giving a $90\%$ increase in throughput on resource-constrained devices~\symbolsecref{sec:results}.
    }
    \label{fig:teaser}
\end{figure}

Pre-trained language-models (PLMs)~\cite{radfordimproving,devlin-etal-2019-bert} have achieved impressive performance on a wide range of natural language processing (NLP) tasks, leading to deployment in the real world~\cite{bommasani2021opportunities}.
Users typically adapt these large models by fine-tuning a separate copy for each task, making storage prohibitively expensive as the number of tasks grows.
Parameter-efficient (PE) methods~\cite{houlsby2019parameter} solve this issue by fine-tuning only a small fraction of model parameters, thus allowing re-use of the PLM backbone which leads to a reduction in storage space ($\approx 90\%$).
While these works have tailored their architectures and performance towards GPUs, an increased adoption of PLMs in resource-constrained devices like mobile phones~\cite{de2020intelligent} requires PE methods that can run on the edge.
In this work, we introduce~\spartan{}, which uses a sparse hierarchical memory to provide a storage and computationally efficient architecture, as illustrated in Figure~\ref{fig:teaser}.

\spartan{} is motivated by cognitive science studies, which posit that information and functional states are sparsely and hierarchically organized in human memory~\cite{mishkin1997hierarchical,hasson2015hierarchical,ahmad2015properties}.
\spartan{} adds a hierarchical memory module after each Transformer layer.
During fine-tuning, it freezes the PLM parameters and adapts its memory via backpropagation to optimize the task loss.
\spartan{}'s memory is organized in two layers containing parent (purple) and children (blue) cells.
The input chooses the top-$K$ parent cells by inducing an attention map over them, and their corresponding children are used to compute and aggregate an output representation which is added to the input.
\spartan{}'s sparse parent selection thus allows it to ignore irrelevant children parameters and makes it computationally efficient~\symbolsecref{sec:results}.

On the GLUE benchmark~\cite{wang2018glue}, \spartan{} performs $0.1$ points better than PE baselines (adapters), while being $90\%$ faster on \raspifull{} (throughput).
Furthermore, in a few-shot setting, \spartan{} can be fine-tuned $34\%$ faster while performing within $0.9$ points of baselines.
Qualitatively, on a news classification dataset (example labels: \textit{entertainment}, \textit{sports}), we observe that \spartan{} distributes responsibility among parent cells by specializing them in different topics (Figure~\ref{fig:qualitative}).
We believe that \spartan{}'s strong performance and speed coupled with its qualitative interpretability can improve the adoption of PE methods on the edge.

\section{Related Work}
\label{sec:related}





\paragraph{Parameter-efficient architectures}
Parameter-efficient (PE) architectures minimize the number of trainable parameters to improve storage efficiency.
\citet{houlsby2019parameter} proposed adapters, which add two feed-forward bottleneck layers (an Adapter) after each Transformer layer while freezing the rest of the model.
Other works have optimized this architecture by experimenting with the placement-order of different components~\cite{pfeiffer2020adapterhub,stickland2019bert,karimi-mahabadi-etal-2021-parameter,ruckle2021adapterdrop,ding2022delta}.
Another line of work fine-tunes a subset of the model's parameters~\cite{zhao2020masking,lee2019would,zaken2022bitfit,guo2021parameter}, and as a result, are typically architecture-dependent, whereas \spartan{} works with all Transformers.
Prompting LMs~\cite{gao2021making,hu2022knowledgeable,li2021prefix} is another popular paradigm, but it involves construction of task-specific templates and is used for smaller datasets~\cite{le2021many}, whereas~\spartan{} is task-agnostic and works for any dataset size.

\paragraph{Memory networks}
Prior works~\cite{weston2015memory,miller2016key,wizard2019memory} have explored the usage of memory in language models.
However, the flat structure of memory makes them computationally expensive.~\citet{chandar2016hierarchical} propose hierarchically-organized memory which uses approximate KNN to improve computation speed, with several applications adapting it~\cite{andrychowicz2016learning,lu2020chime,chen2018hierarchical}.
But to the best of our knowledge, \spartan{} is the first PE architecture for Transformers with sparse hierarchical memory.

\paragraph{NLP on the edge}
NLP methods are increasingly being adopted in mobile and IoT devices~\cite{de2020intelligent,sun2020mobilebert,guo2022efficient,chen2019deep}, and can have lower latency than methods deployed on the cloud~\cite{cartas2019reality,tambe2021edgebert}.
With the introduction of federated learning~\cite{mcmahan2016federated}, where participating devices like mobile phones compute and provide updates to a central model, computing on the edge has become important~\cite{yang2018applied,ramaswamy2019federated,stremmel2021pretraining,liu2021federated}. 
We believe that \spartan{} is an important step in the direction of PE architectures for such devices.
\spartan{} is also loosely related to mixture-of-experts (MoE) architectures~\cite{aljundi2017expert,outrageous2017,gshard2021,du2022glam,multisource2020,fedus2022switch,zoph2022sparse,jacobs1991adaptive}.
But unlike \spartan{}, MoE methods are not parameter-efficient, because all their parameters are trained or fine-tuned.
This significantly increases the storage space of MoE on device, making it less preferable than \spartan{}.
\section{Methodology}
\label{sec:methodology}
\label{sec:methodology:module}

\spartan{} is a parameter-efficient architecture for pre-trained Transformers~\cite{vaswani2017attention} with a sparse, hierarchically organized memory added after each Transformer layer (see Figure~\ref{fig:teaser}).
\spartan{} draws inspiration from cognitive science studies which argue that human memory is sparsely and hierarchically arranged~\cite{mishkin1997hierarchical}.
During fine-tuning, the parameters of the Transformer backbone are frozen and can be re-used, while memory cells are written through gradient updates.
The hierarchical memory contains parent cells and children cells in the first and second levels, respectively.
Each parent cell has multiple exclusive children cells associated with it.
Intuitively, \spartan{} first chooses a sparse subset of parent cells conditioned on the input, and uses the children corresponding to the chosen parent cells to compute an output representation that is added back to the input through a residual connection.
Each position in the input sequence shares the memory.
We provide a mathematical description using the following notation:
Let \inputvector{} $\in \mathbb{R}^d$ be the input to the module, \numparents{} the number of parent cells, and \numchild{} the number of children cells associated with each parent.
Let the stacked parent cells be the matrix $P \in \mathbb{R}^{\trm{\numparents{}} \times d}$ and
the stacked children cells corresponding to parent $P_i$ be the matrix $C_i \in \mathbb{R}^{\trm{\numchild{}} \times d}$.

\paragraph{(1) Choosing the relevant parents}
The input (\inputvector{}) is used to select the \topk{} parent cells (\parentind{}) by inducing an attention distribution computed using an inner product, which allows \spartan{} to sparsely select a subset of relevant parent cells.
\begin{align}
    \begin{split}
        &\trm{\parentprob{}} = \trm{softmax}\left( P \trm{\inputvector{}} \right) \\
        &\trm{\parentind{}} = \trm{\topk{}}\left ( \trm{\parentprob{}} \right )
    \end{split}
\end{align}

\paragraph{(2) Computing the children's cell representation}

As shown in Figure~\ref{fig:teaser}, the children cells contain \textit{key} and \textit{value} components~\cite{weston2015memory}, which we denote by $C_i^{K}$ and $C_i^{V}$, respectively, where $i$ is the parent index.
For each chosen parent $P_i$, we calculate a representation using its children cells:
\begin{align}
    \begin{split}
        v_i = C_i^{V} \trm{softmax}\left ( C_i^{K} \trm{\inputvector{}} \right )
    \end{split}
\end{align}

\paragraph{(3) Hierarchical aggregation}
\spartan{} now combines the children representations by weighting and aggregating them based on the corresponding parent's attention (\parentprob{}).
Since only $K$ parents are chosen, \parentprob{} is re-normalized after ignoring parents not selected in the first stage.
The aggregated output is added back to the input using a residual connection~\cite{he2016deep} and serves as the input to the next layer:
\begin{align}
    \begin{split}
        &Z = \sum_{i \in \trm{\parentind{}}} \trm{\parentprob{}}[i] \\
        &\trm{\outputvector{}} = \frac{1}{Z} \sum_{i \in \trm{\parentind{}}} v_i \cdot \trm{\parentprob{}}[i] \\
        &\trm{\spartan{} output} = \trm{\inputvector{}} + \trm{\outputvector{}}\\
    \end{split}
\end{align}

\section{Experimental Setup}

\begin{table*}[t]
\centering
\resizebox{2\columnwidth}{!}{%
\begin{tabular}{@{}lccccccccccccc@{}}
\toprule
\multirow{2}{*}{\textbf{Model}} &
\multirow{2}{*}{\textbf{Storage ($\downarrow$)}} &
\multicolumn{2}{c}{\textbf{Throughput ($\uparrow$)}} &
\multicolumn{10}{c}{\textbf{GLUE Performance}} \\ \cmidrule(lr){3-4} \cmidrule(lr){5-14}
&
&
\raspi{} &
\iphone{} &
\textbf{Avg.} &
\grayit{\cola{}} &
\grayit{\sst{}} &
\grayit{\mrpc{}} &
\grayit{\qqp{}} &
\grayit{\stsb{}} &
\grayit{\mnli{}$_{\textrm{m/mm}}$} &
\grayit{\qnli{}} &
\grayit{\rte{}} &
\grayit{\wnli{}} \\ \midrule
\textbf{\roberta{}}  & \robertastorage{} & 207.6 & 366.2 & 80.9 & \grayit{\textbf{60.5}} & \grayit{94.3} & \grayit{88.2} & \grayit{91.3}          & \grayit{\textbf{90.7}} & \grayit{\textbf{87.5} / \textbf{87.3}} & \grayit{92.4}          & \grayit{75.3}          & \grayit{47.9} \\ \cdashlinelr{1-14}
\textbf{\pfeiffer{}} & \textbf{\matstorage{}} & 20.0 & 216.1 & 80.9 & \grayit{59.7}          & \grayit{94.2} & \grayit{88.0} & \grayit{89.5}          & \grayit{90.3}          & \grayit{86.8 / 86.9}          & \grayit{92.4}          & \grayit{\textbf{76.8}} & \grayit{50.7} \\
\textbf{\houlsby{}}  & \houlsbystorage{} & 19.5 & 204.8 & 81.0 & \grayit{59.1}          & \grayit{94.3} & \grayit{86.9} & \grayit{\textbf{89.9}} & \grayit{90.5}          & \grayit{87.1 / 87.2}          & \grayit{\textbf{92.6}} & \grayit{\textbf{76.8}} & \grayit{51.6} \\
\textbf{\spartan{}} & \textbf{\matstorage{}} & \textbf{201.3} & \textbf{332.6} & \textbf{81.1} & \grayit{\textbf{60.5}} & \grayit{\textbf{94.4}} & \grayit{\textbf{89.2}} & \grayit{89.6} & \grayit{90.3} & \grayit{86.5 / 86.5} & \grayit{91.9} & \grayit{75.0} & \grayit{\textbf{52.1}} \\
\bottomrule
\end{tabular}
}
\caption{
Full fine-tuning performance on all datasets in the GLUE benchmark.
\spartan{} has the best performance on GLUE ($0.1$ point improvement),
while being $10\times$ faster than parameter-efficient baselines,
and using $87\%$ less storage space when compared to \roberta{}.
All numbers are averaged over three random seeds.
Individual dataset scores are displayed in gray to improve readability.
We provide details about storage computation in Appendix~\ref{app:storage}.
}
\label{tab:main_table}
\end{table*}
\begin{table}[t]
\centering
\resizebox{\columnwidth}{!}{%
\begin{tabular}{@{}lccc@{}}
\toprule
\textbf{Model}       & \textbf{Storage ($\downarrow$)} & \textbf{Fine-tune Throughput ($\uparrow$)} & \textbf{Avg. GLUE} \\ \midrule
\textbf{\roberta{}}  &     \robertastorage{}    &  90.1  &    63.3      \\ \cdashlinelr{1-4}
\textbf{\pfeiffer{}} &     \textbf{\matstorage{}} &   32.1   &   \textbf{64.8}      \\
\textbf{\houlsby{}}  &     \houlsbystorage{} &   34.7   &   63.7      \\
\textbf{\spartan{}}      &     \textbf{\matstorage{}} &   \textbf{53.3}   &   63.9      \\
\bottomrule
\end{tabular}
}
\caption{
    Few-shot results ($200$ instances) on GLUE.
    \spartan{} provides the best storage-throughput-performance trade-off, with $1.5\times$ faster fine-tuning throughput when compared to \pfeiffer{} and \houlsby{}, and significant storage savings when compared to \roberta{}.
    All results are averaged over $3$ seeds.
    Implementation details are presented in Appendix~\ref{app:few_shot}.
}
\label{tab:few_shot}
\end{table}

\paragraph{Datasets}
We use the nine datasets from GLUE~\cite{wang2018glue}, which are CoLA~\cite{colawarstadt2019neural}, SST-2~\cite{sstsocher2013recursive}, MRPC~\cite{mrpcdolan2005automatically}, QQP~\cite{qqp}, STS-B~\cite{stsbcer2017semeval}, MNLI~\cite{mnliwilliams2018broad}, QNLI~\cite{wang2018glue}, RTE~\cite{wang2018glue}, and WNLI~\cite{wnlilevesque2012winograd}.
We use the evaluation metrics suggested by~\citet{wang2018glue} for all datasets; the metrics and averaging are described in Appendix~\ref{app:glue}.

\paragraph{Baselines and \spartan{}}
We use \roberta{}~\cite{liu2019roberta} as the backbone for all the models and also as a baseline; for the latter, we fine-tune all its parameters.
We also compare with two strong parameter-efficient baselines which are variants of adapters: \houlsby{}~\cite{houlsby2019parameter} and \pfeiffer{}~\cite{pfeiffer2021adapterfusion}.
\pfeiffer{} and \spartan{} use the same number of added parameters, while \houlsby{} uses twice as many because it has two bottleneck layers.
We provide model training and implementation details in Appendix~\ref{app:model_and_baselines}.

\paragraph{Speed benchmarking}
We benchmark our models on two resource-constrained edge devices, the Raspberry Pi 4 (4 cores, 8GB RAM) and the iPhone 11 Apple A13 Bionic (6 cores, 4GB RAM), by emulating the corresponding hardware~\cite{buchert2010accurate}.
Following~\citet{chen2019deep}, we measure the throughput, which is the number of instances processed per minute during inference.
More details regarding the benchmarking and emulation are provided in Appendix~\ref{app:speed_benchmarking}.

\section{Results}
\label{sec:results}

\definecolor{business}{HTML}{ffbb78}
\definecolor{entertainment}{HTML}{97df8a}
\definecolor{sports}{HTML}{ff9896}
\definecolor{politics}{HTML}{9fd9e5}

\begin{figure}[t]
    \centering
    \includegraphics[width=.9\columnwidth]{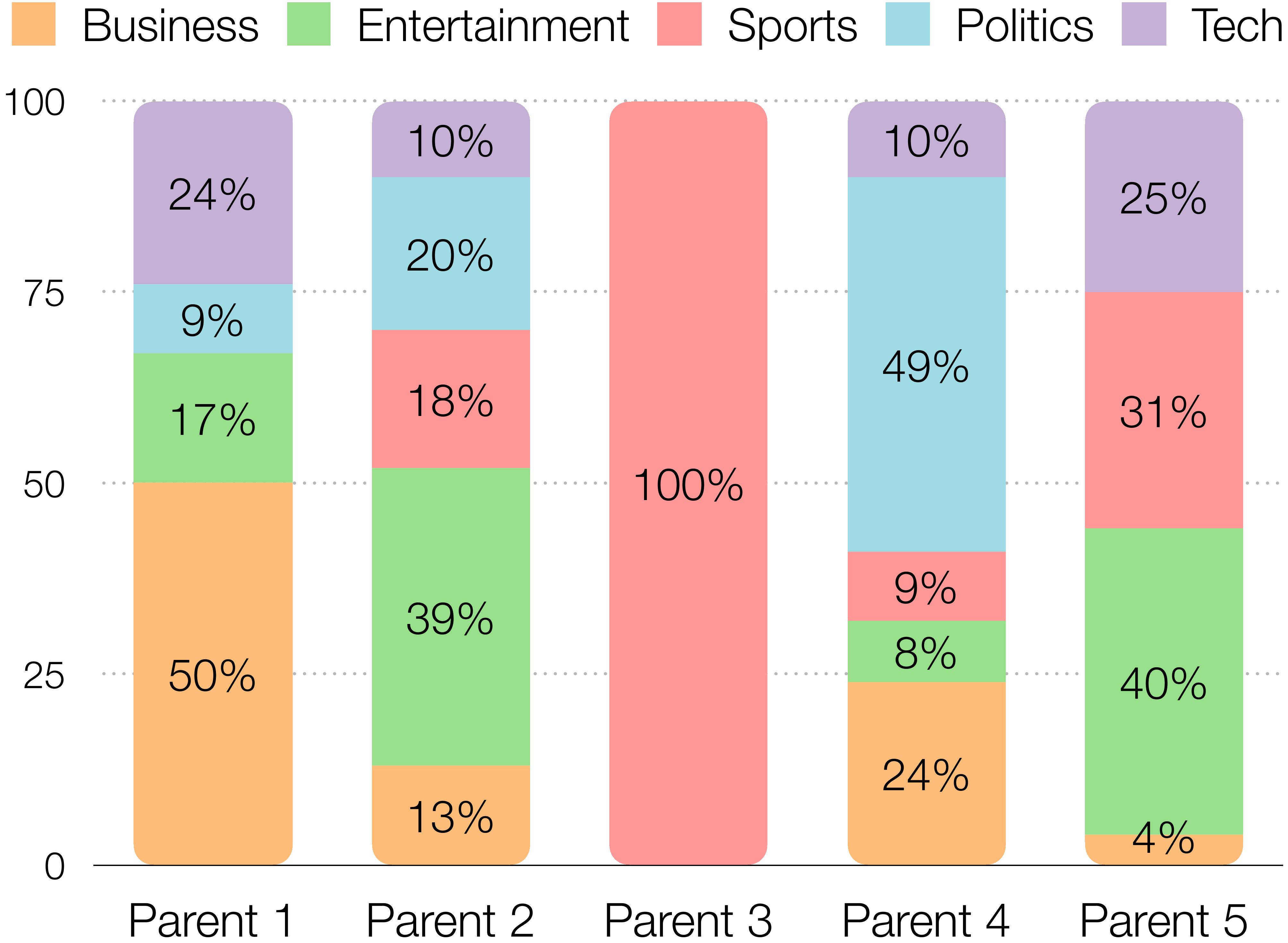}
    \caption{
        Different parent cells specialize in different topics when trained on the BBC news classification dataset.
        Parent $3$ exclusively specializes in \colorbox{sports}{\textit{sports}} while parents $1,2,4$ mainly specialize in \colorbox{business}{\textit{business}}, \colorbox{entertainment}{\textit{entertainment}} and \colorbox{politics}{\textit{politics}}, respectively.
    }
    \label{fig:qualitative}
\end{figure}

\paragraph{Comparing performance and speed on GLUE}
Results in Table~\ref{tab:main_table} show that~\spartan{} is the best performing model with a $0.1$ improvement over~\houlsby{}.
It even outperforms~\roberta{}, which uses $\approx 9 \times$ the storage space and $100 \times$ the number of trainable parameters.
Even though \spartan{} adds memory after each Transformer layer of \roberta{}, its inference throughput is only $3\%$ lower.
\spartan{}'s throughput is significantly higher than \houlsby{} and \pfeiffer{}, with a $10\times$ improvement on \raspifull{} and a $1.6\times$ improvement on \iphone{}.
\spartan{}'s speed-up advantage comes from two factors:
(1) The sparse hierarchical memory ignores children cells corresponding to irrelevant parents, and
(2) it does not use LayerNorm~\cite{ba2016layer}, which can hurt performance on resource-constrained devices~\cite{sun2020mobilebert}.

\paragraph{Few-shot results}
We now consider a few-shot setting, where the model is fine-tuned on resource-constrained devices using just $200$ examples. Table~\ref{tab:few_shot} shows the results.
\spartan{} achieves the best throughput out of all parameter-efficient methods ($1.5\times$) while using only $11\%$ of \roberta{}'s storage space.
Crucially, \spartan{} beats \roberta{} by $0.6$ points, and we posit that it is because of the regularizing effect of having significantly lower trainable parameters, which can be beneficial for the few-shot setting.
\spartan{} gives the best storage-performance-throughput trade-off among all the models.
We note that \spartan{} performs slightly worse than \pfeiffer{} ($0.9$ points), and leave few-shot performance optimization as future work.

\paragraph{Qualitatively analyzing parent cells}
We train \spartan{} on the BBC news classification dataset~\cite{greene2006bbc} and analyze the $5$ parent cells in the last layer.
For each parent, we plot the ground truth label of the instance which picks that parent cell (see Figure~\ref{fig:qualitative}), with implementation details in Appendix~\ref{app:qualitative}.
We notice that parent cells specialize in certain topics, with parent $3$ specializing in \textit{sports} and parents $1,2,4$ specializing \textit{business}, \textit{entertainment} and \textit{politics}, respectively.
This shows that \spartan{}'s strong performance might be due to different parents sharing responsibilities with regard to different topics.

\section{Conclusion}
\label{sec:conclusion}

In this work, we propose \spartan{}, a parameter-efficient architecture that is computationally inexpensive for resource-constrained devices.
\spartan{} uses a two-level sparse and hierarchically-organized memory which allows it to choose only relevant parents and hence parameters, thus speeding up computation.
We believe \spartan{}'s strong performance, which can be coupled with orthogonal methods like pruning~\cite{voita2019analyzing} and distillation~\cite{hinton2015distilling}, makes it a useful architecture for edge devices.
From a qualitative perspective, we find that \spartan{} allows different parents to specialize in different topics, thus ensuring a distribution of responsibility between different groups of parameters.
We believe that these properties can pave the way for more interpretable parameter-efficient properties in the future.

\bibliography{custom}

\begin{thebibliography}{64}
\expandafter\ifx\csname natexlab\endcsname\relax\def\natexlab#1{#1}\fi

\bibitem[{qqp()}]{qqp}

\newblock Quora.
\newblock \url{data.quora.com/First-Quora-Dataset-Release-Question-Pairs}.
\newblock Accessed: 2022-10-15.

\bibitem[{Ahmad and Hawkins(2015)}]{ahmad2015properties}
Subutai Ahmad and Jeff Hawkins. 2015.
\newblock Properties of sparse distributed representations and their
  application to hierarchical temporal memory.
\newblock \emph{arXiv preprint arXiv:1503.07469}.

\bibitem[{Aljundi et~al.(2017)Aljundi, Chakravarty, and
  Tuytelaars}]{aljundi2017expert}
Rahaf Aljundi, Punarjay Chakravarty, and Tinne Tuytelaars. 2017.
\newblock Expert gate: Lifelong learning with a network of experts.
\newblock In \emph{Proceedings of the IEEE Conference on Computer Vision and
  Pattern Recognition}, pages 3366--3375.

\bibitem[{Andrychowicz and Kurach(2016)}]{andrychowicz2016learning}
Marcin Andrychowicz and Karol Kurach. 2016.
\newblock Learning efficient algorithms with hierarchical attentive memory.
\newblock \emph{arXiv preprint arXiv:1602.03218}.

\bibitem[{Ba et~al.(2016)Ba, Kiros, and Hinton}]{ba2016layer}
Jimmy~Lei Ba, Jamie~Ryan Kiros, and Geoffrey~E Hinton. 2016.
\newblock Layer normalization.
\newblock \emph{arXiv preprint arXiv:1607.06450}.

\bibitem[{Bommasani et~al.(2021)Bommasani, Hudson, Adeli, Altman, Arora, von
  Arx, Bernstein, Bohg, Bosselut, Brunskill
  et~al.}]{bommasani2021opportunities}
Rishi Bommasani, Drew~A Hudson, Ehsan Adeli, Russ Altman, Simran Arora, Sydney
  von Arx, Michael~S Bernstein, Jeannette Bohg, Antoine Bosselut, Emma
  Brunskill, et~al. 2021.
\newblock On the opportunities and risks of foundation models.
\newblock \emph{arXiv preprint arXiv:2108.07258}.

\bibitem[{Buchert et~al.(2010)Buchert, Nussbaum, and
  Gustedt}]{buchert2010accurate}
Tomasz Buchert, Lucas Nussbaum, and Jens Gustedt. 2010.
\newblock Accurate emulation of cpu performance.
\newblock In \emph{European Conference on Parallel Processing}, pages 5--12.
  Springer.

\bibitem[{Cartas et~al.(2019)Cartas, Kocour, Raman, Leontiadis, Luque, Sastry,
  Nu{\~n}ez-Martinez, Perino, and Segura}]{cartas2019reality}
Alejandro Cartas, Martin Kocour, Aravindh Raman, Ilias Leontiadis, Jordi Luque,
  Nishanth Sastry, Jose Nu{\~n}ez-Martinez, Diego Perino, and Carlos Segura.
  2019.
\newblock A reality check on inference at mobile networks edge.
\newblock In \emph{Proceedings of the 2nd International Workshop on Edge
  Systems, Analytics and Networking}, pages 54--59.

\bibitem[{Cer et~al.(2017)Cer, Diab, Agirre, Lopez-Gazpio, and
  Specia}]{stsbcer2017semeval}
Daniel Cer, Mona Diab, Eneko Agirre, I{\~n}igo Lopez-Gazpio, and Lucia Specia.
  2017.
\newblock Semeval-2017 task 1: Semantic textual similarity multilingual and
  crosslingual focused evaluation.
\newblock In \emph{Proceedings of the 11th International Workshop on Semantic
  Evaluation (SemEval-2017)}, pages 1--14.

\bibitem[{Chandar et~al.(2016)Chandar, Ahn, Larochelle, Vincent, Tesauro, and
  Bengio}]{chandar2016hierarchical}
Sarath Chandar, Sungjin Ahn, Hugo Larochelle, Pascal Vincent, Gerald Tesauro,
  and Yoshua Bengio. 2016.
\newblock Hierarchical memory networks.
\newblock \emph{arXiv preprint arXiv:1605.07427}.

\bibitem[{Chen et~al.(2018)Chen, Ren, Tang, Zhao, and
  Yin}]{chen2018hierarchical}
Hongshen Chen, Zhaochun Ren, Jiliang Tang, Yihong~Eric Zhao, and Dawei Yin.
  2018.
\newblock Hierarchical variational memory network for dialogue generation.
\newblock In \emph{Proceedings of the 2018 World Wide Web Conference}, pages
  1653--1662.

\bibitem[{Chen and Ran(2019)}]{chen2019deep}
Jiasi Chen and Xukan Ran. 2019.
\newblock Deep learning with edge computing: A review.
\newblock \emph{Proceedings of the IEEE}, 107(8):1655--1674.

\bibitem[{de~Barcelos~Silva et~al.(2020)de~Barcelos~Silva, Gomes, da~Costa,
  da~Rosa~Righi, Barbosa, Pessin, De~Doncker, and
  Federizzi}]{de2020intelligent}
Allan de~Barcelos~Silva, Marcio~Miguel Gomes, Cristiano~Andr{\'e} da~Costa,
  Rodrigo da~Rosa~Righi, Jorge Luis~Victoria Barbosa, Gustavo Pessin, Geert
  De~Doncker, and Gustavo Federizzi. 2020.
\newblock Intelligent personal assistants: A systematic literature review.
\newblock \emph{Expert Systems with Applications}, 147:113193.

\bibitem[{Devlin et~al.(2019)Devlin, Chang, Lee, and
  Toutanova}]{devlin-etal-2019-bert}
Jacob Devlin, Ming-Wei Chang, Kenton Lee, and Kristina Toutanova. 2019.
\newblock \href {https://doi.org/10.18653/v1/N19-1423} {{BERT}: Pre-training of
  deep bidirectional transformers for language understanding}.
\newblock In \emph{Proceedings of the 2019 Conference of the North {A}merican
  Chapter of the Association for Computational Linguistics: Human Language
  Technologies, Volume 1 (Long and Short Papers)}, pages 4171--4186,
  Minneapolis, Minnesota. Association for Computational Linguistics.

\bibitem[{Dinan et~al.(2019)Dinan, Roller, Shuster, Fan, Auli, and
  Weston}]{wizard2019memory}
Emily Dinan, Stephen Roller, Kurt Shuster, Angela Fan, Michael Auli, and Jason
  Weston. 2019.
\newblock \href {https://openreview.net/forum?id=r1l73iRqKm} {Wizard of
  wikipedia: Knowledge-powered conversational agents}.
\newblock In \emph{7th International Conference on Learning Representations,
  {ICLR} 2019, New Orleans, LA, USA, May 6-9, 2019}. OpenReview.net.

\bibitem[{Ding et~al.(2022)Ding, Qin, Yang, Wei, Yang, Su, Hu, Chen, Chan, Chen
  et~al.}]{ding2022delta}
Ning Ding, Yujia Qin, Guang Yang, Fuchao Wei, Zonghan Yang, Yusheng Su,
  Shengding Hu, Yulin Chen, Chi-Min Chan, Weize Chen, et~al. 2022.
\newblock Delta tuning: A comprehensive study of parameter efficient methods
  for pre-trained language models.
\newblock \emph{arXiv preprint arXiv:2203.06904}.

\bibitem[{Dolan and Brockett(2005)}]{mrpcdolan2005automatically}
William~B Dolan and Chris Brockett. 2005.
\newblock Automatically constructing a corpus of sentential paraphrases.
\newblock In \emph{Proceedings of the Third International Workshop on
  Paraphrasing (IWP2005)}.

\bibitem[{Du et~al.(2022)Du, Huang, Dai, Tong, Lepikhin, Xu, Krikun, Zhou, Yu,
  Firat et~al.}]{du2022glam}
Nan Du, Yanping Huang, Andrew~M Dai, Simon Tong, Dmitry Lepikhin, Yuanzhong Xu,
  Maxim Krikun, Yanqi Zhou, Adams~Wei Yu, Orhan Firat, et~al. 2022.
\newblock Glam: Efficient scaling of language models with mixture-of-experts.
\newblock In \emph{International Conference on Machine Learning}, pages
  5547--5569. PMLR.

\bibitem[{Fedus et~al.(2022)Fedus, Zoph, and Shazeer}]{fedus2022switch}
William Fedus, Barret Zoph, and Noam Shazeer. 2022.
\newblock Switch transformers: Scaling to trillion parameter models with simple
  and efficient sparsity.
\newblock \emph{Journal of Machine Learning Research}, 23(120):1--39.

\bibitem[{Gao et~al.(2021)Gao, Fisch, and Chen}]{gao2021making}
Tianyu Gao, Adam Fisch, and Danqi Chen. 2021.
\newblock Making pre-trained language models better few-shot learners.
\newblock In \emph{Proceedings of the 59th Annual Meeting of the Association
  for Computational Linguistics and the 11th International Joint Conference on
  Natural Language Processing (Volume 1: Long Papers)}, pages 3816--3830.

\bibitem[{Greene and Cunningham(2006)}]{greene2006bbc}
Derek Greene and Padraig Cunningham. 2006.
\newblock \href {https://doi.org/10.1145/1143844.1143892} {Practical solutions
  to the problem of diagonal dominance in kernel document clustering}.
\newblock In \emph{Machine Learning, Proceedings of the Twenty-Third
  International Conference {(ICML} 2006), Pittsburgh, Pennsylvania, USA, June
  25-29, 2006}, volume 148 of \emph{{ACM} International Conference Proceeding
  Series}, pages 377--384. {ACM}.

\bibitem[{Guo et~al.(2021)Guo, Rush, and Kim}]{guo2021parameter}
Demi Guo, Alexander~M Rush, and Yoon Kim. 2021.
\newblock Parameter-efficient transfer learning with diff pruning.
\newblock In \emph{Proceedings of the 59th Annual Meeting of the Association
  for Computational Linguistics and the 11th International Joint Conference on
  Natural Language Processing (Volume 1: Long Papers)}, pages 4884--4896.

\bibitem[{Guo et~al.(2022)Guo, Choe, and Lin}]{guo2022efficient}
Liwei Guo, Wonkyo Choe, and Felix~Xiaozhu Lin. 2022.
\newblock \href {https://doi.org/10.48550/arXiv.2207.05022} {Efficient {NLP}
  inference at the edge via elastic pipelining}.
\newblock \emph{CoRR}, abs/2207.05022.

\bibitem[{Hasson et~al.(2015)Hasson, Chen, and Honey}]{hasson2015hierarchical}
Uri Hasson, Janice Chen, and Christopher~J Honey. 2015.
\newblock Hierarchical process memory: memory as an integral component of
  information processing.
\newblock \emph{Trends in cognitive sciences}, 19(6):304--313.

\bibitem[{He et~al.(2016)He, Zhang, Ren, and Sun}]{he2016deep}
Kaiming He, Xiangyu Zhang, Shaoqing Ren, and Jian Sun. 2016.
\newblock Deep residual learning for image recognition.
\newblock In \emph{Proceedings of the IEEE conference on computer vision and
  pattern recognition}, pages 770--778.

\bibitem[{Hinton et~al.(2015)Hinton, Vinyals, and Dean}]{hinton2015distilling}
Geoffrey Hinton, Oriol Vinyals, and Jeff Dean. 2015.
\newblock Distilling the knowledge in a neural network.
\newblock \emph{arXiv preprint arXiv:1503.02531}.

\bibitem[{Houlsby et~al.(2019)Houlsby, Giurgiu, Jastrzebski, Morrone,
  De~Laroussilhe, Gesmundo, Attariyan, and Gelly}]{houlsby2019parameter}
Neil Houlsby, Andrei Giurgiu, Stanislaw Jastrzebski, Bruna Morrone, Quentin
  De~Laroussilhe, Andrea Gesmundo, Mona Attariyan, and Sylvain Gelly. 2019.
\newblock Parameter-efficient transfer learning for nlp.
\newblock In \emph{International Conference on Machine Learning}, pages
  2790--2799. PMLR.

\bibitem[{Hu et~al.(2022)Hu, Ding, Wang, Liu, Wang, Li, Wu, and
  Sun}]{hu2022knowledgeable}
Shengding Hu, Ning Ding, Huadong Wang, Zhiyuan Liu, Jingang Wang, Juanzi Li,
  Wei Wu, and Maosong Sun. 2022.
\newblock Knowledgeable prompt-tuning: Incorporating knowledge into prompt
  verbalizer for text classification.
\newblock In \emph{Proceedings of the 60th Annual Meeting of the Association
  for Computational Linguistics (Volume 1: Long Papers)}, pages 2225--2240.

\bibitem[{Jacobs et~al.(1991)Jacobs, Jordan, Nowlan, and
  Hinton}]{jacobs1991adaptive}
RA~Jacobs, MI~Jordan, SJ~Nowlan, and GE~Hinton. 1991.
\newblock Adaptive mixtures of local experts.
\newblock \emph{Neural Computation}, 3(1):79--87.

\bibitem[{Karimi~Mahabadi et~al.(2021)Karimi~Mahabadi, Ruder, Dehghani, and
  Henderson}]{karimi-mahabadi-etal-2021-parameter}
Rabeeh Karimi~Mahabadi, Sebastian Ruder, Mostafa Dehghani, and James Henderson.
  2021.
\newblock \href {https://doi.org/10.18653/v1/2021.acl-long.47}
  {Parameter-efficient multi-task fine-tuning for transformers via shared
  hypernetworks}.
\newblock In \emph{Proceedings of the 59th Annual Meeting of the Association
  for Computational Linguistics and the 11th International Joint Conference on
  Natural Language Processing (Volume 1: Long Papers)}, pages 565--576, Online.
  Association for Computational Linguistics.

\bibitem[{Le~Scao and Rush(2021)}]{le2021many}
Teven Le~Scao and Alexander~M Rush. 2021.
\newblock How many data points is a prompt worth?
\newblock In \emph{Proceedings of the 2021 Conference of the North American
  Chapter of the Association for Computational Linguistics: Human Language
  Technologies}, pages 2627--2636.

\bibitem[{Lee et~al.(2019)Lee, Tang, and Lin}]{lee2019would}
Jaejun Lee, Raphael Tang, and Jimmy Lin. 2019.
\newblock What would elsa do? freezing layers during transformer fine-tuning.
\newblock \emph{arXiv preprint arXiv:1911.03090}.

\bibitem[{Lepikhin et~al.(2021)Lepikhin, Lee, Xu, Chen, Firat, Huang, Krikun,
  Shazeer, and Chen}]{gshard2021}
Dmitry Lepikhin, HyoukJoong Lee, Yuanzhong Xu, Dehao Chen, Orhan Firat, Yanping
  Huang, Maxim Krikun, Noam Shazeer, and Zhifeng Chen. 2021.
\newblock \href {https://openreview.net/forum?id=qrwe7XHTmYb} {Gshard: Scaling
  giant models with conditional computation and automatic sharding}.
\newblock In \emph{9th International Conference on Learning Representations,
  {ICLR} 2021, Virtual Event, Austria, May 3-7, 2021}. OpenReview.net.

\bibitem[{Levesque et~al.(2012)Levesque, Davis, and
  Morgenstern}]{wnlilevesque2012winograd}
Hector Levesque, Ernest Davis, and Leora Morgenstern. 2012.
\newblock The winograd schema challenge.
\newblock In \emph{Thirteenth international conference on the principles of
  knowledge representation and reasoning}.

\bibitem[{Li and Liang(2021)}]{li2021prefix}
Xiang~Lisa Li and Percy Liang. 2021.
\newblock Prefix-tuning: Optimizing continuous prompts for generation.
\newblock In \emph{Proceedings of the 59th Annual Meeting of the Association
  for Computational Linguistics and the 11th International Joint Conference on
  Natural Language Processing (Volume 1: Long Papers)}, pages 4582--4597.

\bibitem[{Liu et~al.(2021)Liu, Ho, Wang, Gao, Jin, and
  Zhang}]{liu2021federated}
Ming Liu, Stella Ho, Mengqi Wang, Longxiang Gao, Yuan Jin, and He~Zhang. 2021.
\newblock Federated learning meets natural language processing: A survey.
\newblock \emph{arXiv preprint arXiv:2107.12603}.

\bibitem[{Liu et~al.(2019)Liu, Ott, Goyal, Du, Joshi, Chen, Levy, Lewis,
  Zettlemoyer, and Stoyanov}]{liu2019roberta}
Yinhan Liu, Myle Ott, Naman Goyal, Jingfei Du, Mandar Joshi, Danqi Chen, Omer
  Levy, Mike Lewis, Luke Zettlemoyer, and Veselin Stoyanov. 2019.
\newblock Roberta: A robustly optimized bert pretraining approach.
\newblock \emph{arXiv preprint arXiv:1907.11692}.

\bibitem[{Lu et~al.(2020)Lu, Pergola, Gui, Li, and He}]{lu2020chime}
Junru Lu, Gabriele Pergola, Lin Gui, Binyang Li, and Yulan He. 2020.
\newblock Chime: Cross-passage hierarchical memory network for generative
  review question answering.
\newblock In \emph{Proceedings of the 28th International Conference on
  Computational Linguistics}, pages 2547--2560.

\bibitem[{McMahan et~al.(2016)McMahan, Moore, Ramage, and
  y~Arcas}]{mcmahan2016federated}
H~Brendan McMahan, Eider Moore, Daniel Ramage, and Blaise~Ag{\"u}era y~Arcas.
  2016.
\newblock Federated learning of deep networks using model averaging.
\newblock \emph{arXiv preprint arXiv:1602.05629}, 2.

\bibitem[{Miller et~al.(2016)Miller, Fisch, Dodge, Karimi, Bordes, and
  Weston}]{miller2016key}
Alexander Miller, Adam Fisch, Jesse Dodge, Amir-Hossein Karimi, Antoine Bordes,
  and Jason Weston. 2016.
\newblock Key-value memory networks for directly reading documents.
\newblock In \emph{Proceedings of the 2016 Conference on Empirical Methods in
  Natural Language Processing}, pages 1400--1409.

\bibitem[{Mishkin et~al.(1997)Mishkin, Suzuki, Gadian, and
  Vargha-Khadem}]{mishkin1997hierarchical}
Mortimer Mishkin, Wendy~A Suzuki, David~G Gadian, and Faraneh Vargha-Khadem.
  1997.
\newblock Hierarchical organization of cognitive memory.
\newblock \emph{Philosophical Transactions of the Royal Society of London.
  Series B: Biological Sciences}, 352(1360):1461--1467.

\bibitem[{Pfeiffer et~al.(2021)Pfeiffer, Kamath, R{\"u}ckl{\'e}, Cho, and
  Gurevych}]{pfeiffer2021adapterfusion}
Jonas Pfeiffer, Aishwarya Kamath, Andreas R{\"u}ckl{\'e}, Kyunghyun Cho, and
  Iryna Gurevych. 2021.
\newblock Adapterfusion: Non-destructive task composition for transfer
  learning.
\newblock In \emph{Proceedings of the 16th Conference of the European Chapter
  of the Association for Computational Linguistics: Main Volume}, pages
  487--503.

\bibitem[{Pfeiffer et~al.(2020)Pfeiffer, R{\"u}ckl{\'e}, Poth, Kamath,
  Vuli{\'c}, Ruder, Cho, and Gurevych}]{pfeiffer2020adapterhub}
Jonas Pfeiffer, Andreas R{\"u}ckl{\'e}, Clifton Poth, Aishwarya Kamath, Ivan
  Vuli{\'c}, Sebastian Ruder, Kyunghyun Cho, and Iryna Gurevych. 2020.
\newblock Adapterhub: A framework for adapting transformers.
\newblock In \emph{Proceedings of the 2020 Conference on Empirical Methods in
  Natural Language Processing: System Demonstrations}, pages 46--54.

\bibitem[{Radford et~al.()Radford, Narasimhan, Salimans, and
  Sutskever}]{radfordimproving}
Alec Radford, Karthik Narasimhan, Tim Salimans, and Ilya Sutskever.
\newblock Improving language understanding by generative pre-training.

\bibitem[{Ramaswamy et~al.(2019)Ramaswamy, Mathews, Rao, and
  Beaufays}]{ramaswamy2019federated}
Swaroop Ramaswamy, Rajiv Mathews, Kanishka Rao, and Fran{\c{c}}oise Beaufays.
  2019.
\newblock Federated learning for emoji prediction in a mobile keyboard.
\newblock \emph{arXiv preprint arXiv:1906.04329}.

\bibitem[{R{\"u}ckl{\'e} et~al.(2021)R{\"u}ckl{\'e}, Geigle, Glockner, Beck,
  Pfeiffer, Reimers, and Gurevych}]{ruckle2021adapterdrop}
Andreas R{\"u}ckl{\'e}, Gregor Geigle, Max Glockner, Tilman Beck, Jonas
  Pfeiffer, Nils Reimers, and Iryna Gurevych. 2021.
\newblock Adapterdrop: On the efficiency of adapters in transformers.
\newblock In \emph{Proceedings of the 2021 Conference on Empirical Methods in
  Natural Language Processing}, pages 7930--7946.

\bibitem[{Shazeer et~al.(2017)Shazeer, Mirhoseini, Maziarz, Davis, Le, Hinton,
  and Dean}]{outrageous2017}
Noam Shazeer, Azalia Mirhoseini, Krzysztof Maziarz, Andy Davis, Quoc~V. Le,
  Geoffrey~E. Hinton, and Jeff Dean. 2017.
\newblock \href {https://openreview.net/forum?id=B1ckMDqlg} {Outrageously large
  neural networks: The sparsely-gated mixture-of-experts layer}.
\newblock In \emph{5th International Conference on Learning Representations,
  {ICLR} 2017, Toulon, France, April 24-26, 2017, Conference Track
  Proceedings}. OpenReview.net.

\bibitem[{Socher et~al.(2013)Socher, Perelygin, Wu, Chuang, Manning, Ng, and
  Potts}]{sstsocher2013recursive}
Richard Socher, Alex Perelygin, Jean Wu, Jason Chuang, Christopher~D Manning,
  Andrew~Y Ng, and Christopher Potts. 2013.
\newblock Recursive deep models for semantic compositionality over a sentiment
  treebank.
\newblock In \emph{Proceedings of the 2013 conference on empirical methods in
  natural language processing}, pages 1631--1642.

\bibitem[{Stickland and Murray(2019)}]{stickland2019bert}
Asa~Cooper Stickland and Iain Murray. 2019.
\newblock Bert and pals: Projected attention layers for efficient adaptation in
  multi-task learning.
\newblock In \emph{International Conference on Machine Learning}, pages
  5986--5995. PMLR.

\bibitem[{Stremmel and Singh(2021)}]{stremmel2021pretraining}
Joel Stremmel and Arjun Singh. 2021.
\newblock Pretraining federated text models for next word prediction.
\newblock In \emph{Future of Information and Communication Conference}, pages
  477--488. Springer.

\bibitem[{Sun et~al.(2020)Sun, Yu, Song, Liu, Yang, and
  Zhou}]{sun2020mobilebert}
Zhiqing Sun, Hongkun Yu, Xiaodan Song, Renjie Liu, Yiming Yang, and Denny Zhou.
  2020.
\newblock Mobilebert: a compact task-agnostic bert for resource-limited
  devices.
\newblock In \emph{Proceedings of the 58th Annual Meeting of the Association
  for Computational Linguistics}, pages 2158--2170.

\bibitem[{Tambe et~al.(2021)Tambe, Hooper, Pentecost, Jia, Yang, Donato, Sanh,
  Whatmough, Rush, Brooks et~al.}]{tambe2021edgebert}
Thierry Tambe, Coleman Hooper, Lillian Pentecost, Tianyu Jia, En-Yu Yang, Marco
  Donato, Victor Sanh, Paul Whatmough, Alexander~M Rush, David Brooks, et~al.
  2021.
\newblock Edgebert: Sentence-level energy optimizations for latency-aware
  multi-task nlp inference.
\newblock In \emph{MICRO-54: 54th Annual IEEE/ACM International Symposium on
  Microarchitecture}, pages 830--844.

\bibitem[{Turc et~al.(2019)Turc, Chang, Lee, and Toutanova}]{turc2019well}
Iulia Turc, Ming-Wei Chang, Kenton Lee, and Kristina Toutanova. 2019.
\newblock Well-read students learn better: On the importance of pre-training
  compact models.
\newblock \emph{arXiv preprint arXiv:1908.08962}.

\bibitem[{Vaswani et~al.(2017)Vaswani, Shazeer, Parmar, Uszkoreit, Jones,
  Gomez, Kaiser, and Polosukhin}]{vaswani2017attention}
Ashish Vaswani, Noam Shazeer, Niki Parmar, Jakob Uszkoreit, Llion Jones,
  Aidan~N Gomez, {\L}ukasz Kaiser, and Illia Polosukhin. 2017.
\newblock Attention is all you need.
\newblock \emph{Advances in neural information processing systems}, 30.

\bibitem[{Voita et~al.(2019)Voita, Talbot, Moiseev, Sennrich, and
  Titov}]{voita2019analyzing}
Elena Voita, David Talbot, Fedor Moiseev, Rico Sennrich, and Ivan Titov. 2019.
\newblock Analyzing multi-head self-attention: Specialized heads do the heavy
  lifting, the rest can be pruned.
\newblock In \emph{Proceedings of the 57th Annual Meeting of the Association
  for Computational Linguistics}, pages 5797--5808.

\bibitem[{Wang et~al.(2018)Wang, Singh, Michael, Hill, Levy, and
  Bowman}]{wang2018glue}
Alex Wang, Amanpreet Singh, Julian Michael, Felix Hill, Omer Levy, and Samuel~R
  Bowman. 2018.
\newblock Glue: A multi-task benchmark and analysis platform for natural
  language understanding.
\newblock In \emph{International Conference on Learning Representations}.

\bibitem[{Warstadt et~al.(2019)Warstadt, Singh, and
  Bowman}]{colawarstadt2019neural}
Alex Warstadt, Amanpreet Singh, and Samuel Bowman. 2019.
\newblock Neural network acceptability judgments.
\newblock \emph{Transactions of the Association for Computational Linguistics},
  7:625--641.

\bibitem[{Weston et~al.(2015)Weston, Chopra, and Bordes}]{weston2015memory}
Jason Weston, Sumit Chopra, and Antoine Bordes. 2015.
\newblock \href {http://arxiv.org/abs/1410.3916} {Memory networks}.
\newblock In \emph{3rd International Conference on Learning Representations,
  {ICLR} 2015, San Diego, CA, USA, May 7-9, 2015, Conference Track
  Proceedings}.

\bibitem[{Williams et~al.(2018)Williams, Nangia, and
  Bowman}]{mnliwilliams2018broad}
Adina Williams, Nikita Nangia, and Samuel~R Bowman. 2018.
\newblock A broad-coverage challenge corpus for sentence understanding through
  inference.
\newblock In \emph{NAACL-HLT}.

\bibitem[{Wright and Augenstein(2020)}]{multisource2020}
Dustin Wright and Isabelle Augenstein. 2020.
\newblock \href {https://doi.org/10.18653/v1/2020.emnlp-main.639} {Transformer
  based multi-source domain adaptation}.
\newblock In \emph{Proceedings of the 2020 Conference on Empirical Methods in
  Natural Language Processing, {EMNLP} 2020, Online, November 16-20, 2020},
  pages 7963--7974. Association for Computational Linguistics.

\bibitem[{Yang et~al.(2018)Yang, Andrew, Eichner, Sun, Li, Kong, Ramage, and
  Beaufays}]{yang2018applied}
Timothy Yang, Galen Andrew, Hubert Eichner, Haicheng Sun, Wei Li, Nicholas
  Kong, Daniel Ramage, and Fran{\c{c}}oise Beaufays. 2018.
\newblock Applied federated learning: Improving google keyboard query
  suggestions.
\newblock \emph{arXiv preprint arXiv:1812.02903}.

\bibitem[{Zaken et~al.(2022)Zaken, Goldberg, and Ravfogel}]{zaken2022bitfit}
Elad~Ben Zaken, Yoav Goldberg, and Shauli Ravfogel. 2022.
\newblock Bitfit: Simple parameter-efficient fine-tuning for transformer-based
  masked language-models.
\newblock In \emph{Proceedings of the 60th Annual Meeting of the Association
  for Computational Linguistics (Volume 2: Short Papers)}, pages 1--9.

\bibitem[{Zhao et~al.(2020)Zhao, Lin, Mi, Jaggi, and
  Sch{\"u}tze}]{zhao2020masking}
Mengjie Zhao, Tao Lin, Fei Mi, Martin Jaggi, and Hinrich Sch{\"u}tze. 2020.
\newblock Masking as an efficient alternative to finetuning for pretrained
  language models.
\newblock In \emph{Proceedings of the 2020 Conference on Empirical Methods in
  Natural Language Processing (EMNLP)}, pages 2226--2241.

\bibitem[{Zoph(2022)}]{zoph2022sparse}
Barret Zoph. 2022.
\newblock \href {https://doi.org/10.1109/IPDPSW55747.2022.00171} {Designing
  effective sparse expert models}.
\newblock In \emph{{IEEE} International Parallel and Distributed Processing
  Symposium, {IPDPS} Workshops 2022, Lyon, France, May 30 - June 3, 2022}, page
  1044. {IEEE}.

\end{thebibliography}
\bibliographystyle{acl_natbib}

\newpage
\appendix

\begin{table*}[t]
\centering
\resizebox{2\columnwidth}{!}{%
\begin{tabular}{@{}lcccccccccccc@{}}
\toprule
\multirow{2}{*}{\textbf{Model}} &
\multirow{2}{*}{\textbf{Storage ($\downarrow$)}} &
\multirow{2}{*}{\textbf{Fine-tune throughput ($\uparrow$)}} &
\multicolumn{10}{c}{\textbf{Performance}} \\ \cmidrule(lr){4-13}
&
&
&
\textbf{Avg.} &
\cola{} &
\sst{} &
\mrpc{} &
\qqp{} &
\stsb{} &
\mnli{} (m/mm) &
\qnli{} &
\rte{} &
\wnli{} \\ \midrule
\textbf{\roberta{}} & \robertastorage{} & 21.3 & 63.3 & 34.6 & 87.3 & 77.7 & 71.8 & 74.4 & \textbf{57.7} / \textbf{61.1} & \textbf{77.8} & \textbf{53.9} & 34.7 \\ \cdashlinelr{1-13}
\textbf{\pfeiffer{}} & \textbf{\matstorage{}} & 55.3 & \textbf{64.8} & \textbf{39.7} & 87.5 & 78.6 & 74.3 & \textbf{79.8} & 54.4 / 58.2 & 76.1 & 52.6 & 39.9 \\
\textbf{\houlsby{}} & \houlsbystorage{} & 59.8 & 63.7 & 37.7 & 87.1 & 77.9 & \textbf{75.1} & 75.1 & 55.2 / 59.1 & 75.8 & 51.0 & 38.0 \\
\textbf{\spartan{}} & \textbf{\matstorage{}} & \textbf{36.1} & 63.9 & 38.3 & \textbf{87.6} & \textbf{78.7} & 71.8 & 79.0 & 52.1 / 55.8 & 75.4 & 49.7 & \textbf{42.3} \\
\bottomrule
\end{tabular}
}
\caption{
    Few shot ($K=200$) full results on all datasets in the GLUE benchmark.
    \spartan{} provides the best storage-throughput-performance trade-off, with significant improvements in fine-tuning throughput when compared to other parameter efficient methods (\pfeiffer{} and \houlsby{}), significant storage savings when compared to \roberta{}.
    All results are averaged over $3$ seeds.
}
\label{app:tab:few_shot}
\end{table*}

\section{Implementation details}
\label{app:model_and_baselines}

We use the \roberta{}-Base model~\cite{liu2019roberta} as the backbone for all the models.
All models are trained on NVIDIA RTX 2080s.
For each dataset and model, we perform a hyperparameter search using $10\%$ of the training data.
We use a batch size of $16$ for all the models.
Across all models, MNLI, QQP, SST-2, and WNLI performed best when fine-tuned on $3$ epochs, and other datasets needed $20$ epochs. We used a grid of $\{3,20\}$ by following~\cite{houlsby2019parameter}.
Further, we find that the following are the best performing hyperparameters for specific baselines and models when we use 10\% of the train dataset to choose them.

\begin{enumerate}
    \item \roberta{}: Learning rate: $2e-5$
    \item \pfeiffer{}: Learning rate: $1e-4$, Reduction factor: $12$ (Bottleneck size: $64$)
    \item \houlsby{}: Learning rate: $1e-4$, Reduction factor: $12$ (Bottleneck size: $64$)
    \item \spartan{}: Learning rate: $1e-3$, Number of parent cells: 16, Number of children cells per parent: 3. The top $K=8$ parents are chosen for children memory computation. \spartan{} adds exactly the same number of trainable parameters as \pfeiffer{} and half that of \houlsby{}. As explained in Section~\ref{sec:methodology:module}, both the parent and children cells are of dimensionality $d=768$, the same as the hidden dimensionality of the base Transformer model they are using.
    \item Learning rate grid search: $\{2e-5, 1e-4, 1e-3 \}$, Reduction factor grid search: $\{12, 16 \}$, Parent cells grid search: $\{12, 16 \}$
\end{enumerate}



\section{Few-shot full results}
\label{app:few_shot}

We present the full version of Table~\ref{tab:few_shot} in Table~\ref{app:tab:few_shot} which includes the breakdown for all the GLUE datasets.
We use $K=200$ examples in the train dataset and following~\citet{gao2021making}, we train all models on $1000$ steps and evaluate it on the same validation dataset.
All other hyperparameters are the same as ones mentioned in Appendix~\ref{app:model_and_baselines}.

\section{GLUE evaluation metrics}
\label{app:glue}

We use accuracy for SST, QQP, MNLI, QNLI, RTE, and WNLI, combined score of pearson and spearman correlation for STS-B, and matthews correlation for CoLA.
When averaging, we use only the MNLI-m score for the MNLI task.
All results are reported on the validation dataset of GLUE.
No hyperparameter tuning was performed on the validation set.

\section{Model and dataset details for qualitative analysis}
\label{app:qualitative}

We use a BERT-small architecture~\cite{turc2019well} and consider \spartan{} model with $5$ parent cells and $1$ child cell corresponding to each parent to make the qualitative analysis transparent.
Thus, choosing a certain parent is equivalent to choosing the corresponding child.
We train the model on the BBC-news classification dataset for $10$ epochs.

\section{Storage details and parameter computation}
\label{app:storage}
For the nine GLUE datasets, \roberta{} uses 4.43 GB of storage space, \houlsby{} uses 0.58 GB, and \pfeiffer{} and \spartan{} use 0.53 GB.
In terms of the number of parameters, \roberta{} uses 1107 million, \houlsby{} uses 144.23 million, and \pfeiffer{} and \spartan{} use 133.62 million.
The formular for \spartan{}'s calculation is the following, where $T$ is the number of tasks, $P$ is the number of parents, $C$ is the number of children, $d$ is the hidden dimensionality, and $L$ is the number of layers, and $N_{\textrm{\roberta}}$ is the number of parameters in the base \roberta{} model.

\begin{align*}
    \begin{split}
        N_{\textrm{\roberta}} + 2 T\times (P + P\times C) \times d  L
    \end{split}
\end{align*}

\section{Speed benchmarking}
\label{app:speed_benchmarking}

To emulate \raspifull{} and \iphone{}, we use the Linux \texttt{cgroups} command to enforce the memory limit.
The number of CPU cores are enforced using the \texttt{slurmctld} command.
All experiments are conducted using the CPU version of PyTorch.
To measure the throughput, we consistently use a batch size of 32.




\section{Limitations}
\label{app:limitations}
Our speed benchmarking results have been thoroughly conducted and emulated on the hardware we have available, but we could not run the experiments on a physical Raspberry Pi and iPhone 11.
We also wish to extend our work to languages other than English.

\section{Risks}
\label{app:risks}
We do not see any potential risks for our architecture and do not release any model weights.
But we will open source our code and have provided it as part of the supplementary material.

\end{document}